\documentclass[10pt,twocolumn,letterpaper]{article}

\usepackage{iccv}
\usepackage{times}
\usepackage{epsfig}
\usepackage{graphicx}
\usepackage{amsmath}
\usepackage{amssymb}

\usepackage{bm}
\usepackage{bbm}
\usepackage{braket}

\usepackage{multirow}

\newcommand{\vect}[1]{\boldsymbol{#1}}
\newcommand{\R}{\mathbb{R}}

\usepackage[pagebackref=true,breaklinks=true,letterpaper=true,colorlinks,bookmarks=false]{hyperref}

\iccvfinalcopy %

\def\iccvPaperID{3430} %

\ificcvfinal\fi

\begin{document}

\title{Optimizing Implicit Neural Representations from Point Clouds via Energy-Based Models}

\author{Ryutaro Yamauchi\thanks{This work was done when the authors were at Albert Inc. Thanks to Moriyuki Oshima for assisting with the experiment.}\\
The University of Tokyo \\ 
{\tt\small ryutaro\_yamauchi@weblab.t.u-tokyo.ac.jp}
\and
Jinya Sakurai\footnotemark[1]\\
KAUST\\
{\tt\small jinya.sakurai@kaust.edu.sa}
\and
Ryo Furukawa\footnotemark[1]\\
{\tt\small rfurukaward@gmail.com}
\and
Tatsushi Matsubayashi\footnotemark[1]\\
{\tt\small tatsushi.matsubayashi@gmail.com}
}

\maketitle
\ificcvfinal\thispagestyle{empty}\fi

\begin{abstract}
    Reconstructing a continuous surface from an unoritented 3D point cloud is a fundamental task in 3D shape processing. In recent years, several methods have been proposed to address this problem using implicit neural representations (INRs). In this study, we propose a method to optimize INRs using energy-based models (EBMs). By employing the absolute value of the coordinate-based neural networks as the energy function, the INR can be optimized through the estimation of the point cloud distribution by the EBM. In addition, appropriate parameter settings of the EBM enable the model to consider the magnitude of point cloud noise. Our experiments confirmed that the proposed method is more robust against point cloud noise than conventional surface reconstruction methods.
\end{abstract}

\section{Introduction}

\begin{figure*}[tbp]
\begin{center}
\includegraphics[width=0.99 \linewidth]{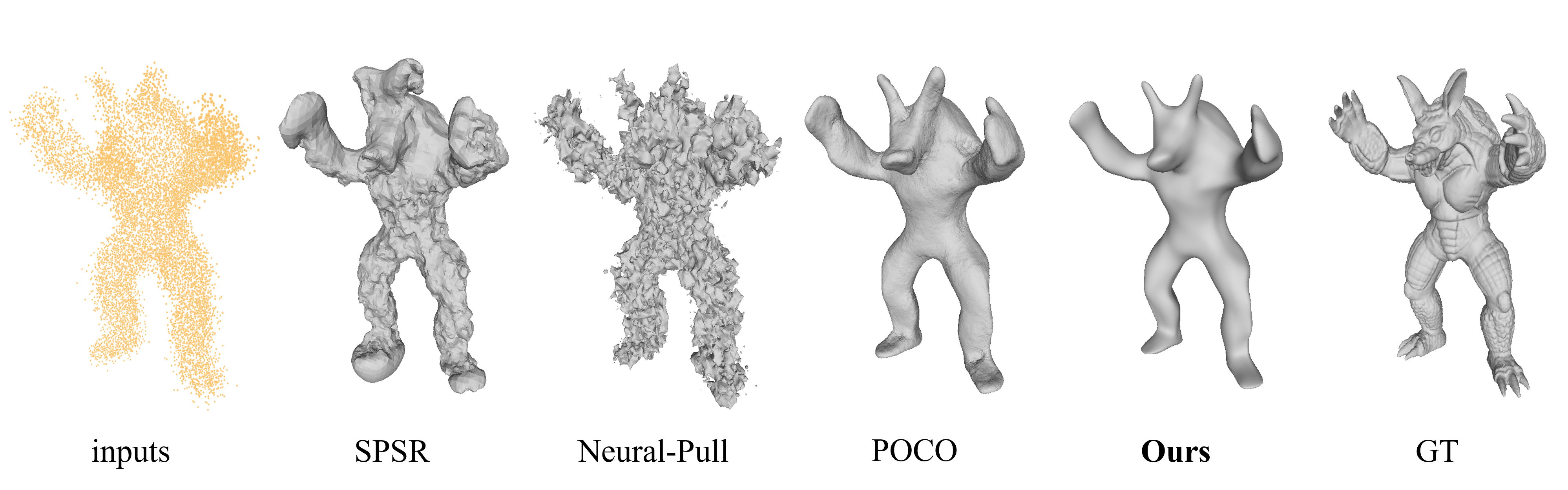}
\end{center}
   \caption{We present a non-data-driven surface reconstruction method that can handle point cloud noise by using EBM. It is more robust to noise than the non-data-driven methods (SPSR~\cite{kazhdan2006poisson}, Neural-Pull~\cite{ma2020neuralpull}) that does not consider point cloud noise, and unlike data-driven methods such as POCO~\cite{boulch2022poco}, it does not depend on training data. These figures are qualitative comparisons of surface reconstruction results for a noise scale $\sigma=0.03$.}
\label{fig:qualitativeshort}
\end{figure*}

Reconstructing a continuous surface from unstructured point clouds is an essential step in 3D shape processing. As high-performance 3D scanners become available at lower cost and the range of applications expands, such as AR/VR, the demand for high-quality surface reconstruction algorithms is increasing, but surface reconstruction algorithms are still limited and are the subject of active research. A surface reconstruction algorithm should be able to handle raw point clouds with noise and variations in sampling density. It should also work in general for a variety of surface topologies and different shapes.

Classically, surface reconstruction has been performed by optimizing models introducing various geometric priors for a given point cloud. Classical surface reconstruction methods are non-data-driven in the sense that they do not require point clouds other than the point cloud for which we want to reconstruct a surface. Some of them are widely used through open source libraries, such as~\cite{bernardini1999ball,kazhdan2013screened} implemented in MeshLab~\cite{cignoni2008meshlab}.

Recently, surface reconstruction algorithms using neural networks (NNs) have been actively studied. In particular, methods to obtain implicit neural representations (INRs) of object surfaces have developed since~\cite{park2019deepsdf,mescheder2019occupancy,chen2019imnet}. In these methods, a surface is obtained as some level-set of an implicit function approximated by coordinated-based NNs, which take spatial coordinates as a part of inputs. As an implicit function, the signed distance function (SDF) as~\cite{park2019deepsdf} or the occupancy field as~\cite{mescheder2019occupancy,chen2019imnet} of an object is usually used. These methods are usually data-driven in the sense that NNs are learned using training data. By training on a dataset containing pairs of corresponding point clouds and surfaces, these methods can perform surface reconstruction that is robust to adverse conditions in the point cloud, such as noise, missing points, and variations in scan density. On the other hand, these data-driven methods may have the weakness that it is challenging to represent surface geometry outside the learned shape space. As a result, they may fail to reproduce the shape details, or the reconstruction may fail for unknown shapes which are far from training data, such as different categories of shapes. In order to achieve more accurate reconstruction and to reduce the dependence on the training data, various methods have been proposed, for example, learning local features and patches~\cite{peng2020convolutional,chibane2020implicit,erler2020Points2Surf,chabra2020deepls,jiang2020lig,tretschk2020patchnets,mi2020ssrnet,ummenhofer2021adaptive,boulch2022poco}. Data-driven methods require training data with ground truth values, which are not readily available in practice.

Some methods using INRs~\cite{atzmon2019controlling,atzmon2020sal,grop2020igr,atzmon2021sald,basher2021lightsal,zhao2021sign,sitzmann2020siren,ma2020neuralpull,wang2021neuralimls,williams2021spline} learn implicit neural representations from raw point clouds, and some of them can be used as non-data-driven methods because they can be used with or without prior training. Non-data-driven methods have the advantage that the surface reconstruction capability is independent of the training data. Instead, they need to compensate for the lack of a priori knowledge of geometry by appropriate loss function design and regularization. %
Although some methods are reported to be robust to noise, it is still challenging to be robust to noise for many non-data-driven methods using INRs. 

In this paper, for non-data-driven and noise-robust surface reconstruction from point clouds, we propose a method incorporating implicit neural representations (INRs) into energy-based models (EBMs). The EBMs are models that learn un-normalized probability distributions~\cite{lecun2006tutorial,song2021train}. In the proposed method, the input point cloud is assumed to be sampled from a continuous probability distribution whose mode is equal to the object surface. Here, each object's surface is assumed to be closed and smooth. We model this point cloud generating distribution by EBM and optimize the energy function by maximum likelihood estimation using Langevin dynamics. When using the absolute value of the coordinate-based NN as the energy function, the probability distribution derived from this energy function has a maximum density at the zero level set of the coordinate-based NN.  As a result, the INR of the shape is optimized through the estimation of the point cloud distribution. 
The temperature parameter $\beta$ of the EBM corresponds to the point cloud noise scale, so adjusting $\beta$ makes it possible to explicitly model the noise scale.

Our contributions are as follows: (1) We proposed a method to optimize INRs using EBMs. (2) We realized surface reconstruction considering point cloud noise without pre-training by matching the $\beta$ of EBMs with the scale of the noise. (3) Comparison experiments with various previous studies were conducted.

\section{Related Works}
\label{relatedworks}
\subsection{Surface Reconstruction}
{\bf{Classical methods}} – The Poisson Surface Reconstruction (PSR)~\cite{kazhdan2006poisson} and its variants~\cite{kazhdan2013screened,kazhdan2020poissonwenv} estimate an indicator function based on the Poisson equation. Screened Poisson Surface Reconstruction (SPSR)~\cite{kazhdan2013screened} is currently a widely used method and is implemented in MeshLab~\cite{cignoni2008meshlab}, an open-source system for processing and editing 3D triangular meshes.

{\bf{Implicit neural representations from ground truth data}} – Data-driven methods to obtain implicit neural representations (INRs) of object surfaces have developed since~\cite{park2019deepsdf,mescheder2019occupancy,chen2019imnet}, all of which use only global features. 
Later, various methods learning local features and patches have been proposed~\cite{peng2020convolutional,chibane2020implicit,erler2020Points2Surf,chabra2020deepls,jiang2020lig,tretschk2020patchnets,mi2020ssrnet,ummenhofer2021adaptive,boulch2022poco}. 
Points2Surf~\cite{erler2020Points2Surf} learns combinations of detailed local patches for distances and coarse global information for signs to improve generalization performance. POCO~\cite{boulch2022poco} is a method that excels in performance and scalability, building the latent features at each input point by convolution and estimating the occupancy of any query point by aggregating the features of neighboring points with attentive decoding.

{\bf{Implicit neural representations from raw scans}} – Some methods using INRs~\cite{atzmon2019controlling,atzmon2020sal,atzmon2021sald,basher2021lightsal,zhao2021sign,grop2020igr,sitzmann2020siren,ma2020neuralpull,wang2021neuralimls,williams2021spline} learn implicit neural representations from raw scans including raw point clouds and they can be used as non-data-driven methods. Non-data-driven methods have the reconstruction capability, which is independent of the training data, but usually are not robust to noise. 
Sign Agnostic Learning (SAL)~\cite{atzmon2020sal} can reconstruct a surface from an un-oriented point cloud, and it is developed in~\cite{atzmon2021sald,basher2021lightsal,zhao2021sign}. IGR~\cite{grop2020igr} reconstructs a surface using a simple loss function including the {\it Eikonal term} which encourages the norm of the gradient $\nabla_{\bm{x}} f_\theta(\bm{x})$ of the implicit NN $f_\theta$ to be equal to one. SIREN~\cite{sitzmann2020siren} proposed a new activation function for implicit neural representation for complex signals. Neural-Pull~\cite{ma2020neuralpull} optimize NNs by pulling query points to their closest points by using values and gradients of the predicted SDF. Neural Splines~\cite{williams2021spline} reconstructs a surface based on random feature kernels arising from infinitely-wide shallow ReLU networks. NeuralIMLS~\cite{wang2021neuralimls} learns the SDF robust to noise in a self-supervised fashion using the implicit moving least-square function (IMLS) but does not incorporate noise to model explicitly.

In this research, we propose a non-data-driven surface reconstruction method from point clouds using an implicit neural representation to be independent of training data.  To obtain robustness to noise, we incorporate this implicit neural representation as an energy function of the EBM. In the objective function for optimizing EBM, we use a regularization term of the same form as the Eikonal term.

\subsection{Energy-Based Models and Shape Representation}
Energy-based models (EBMs) model the distribution of data as a probability distribution (Gibbs distribution) using an energy function. Direct calculation of the normalization constant (the partition function) is avoided in the modeling~\cite{lecun2006tutorial}. Training methods for EBMs include likelihood maximization, score matching and noise contrastive estimation~\cite{song2021train}. Various functions can be used as the energy function, and its high degree of freedom makes it suitable for various applications such as image generation. \cite{du2019implicitEBM} shows that EBMs using Markov chain Monte Carlo (MCMC) sampling for training NNs scale to high dimensional data such as images. EBMs are used for shape representation, such as point cloud generation~\cite{xie2021generative, cai2020learning} and voxel shape generation~\cite{xie2020gvoxelnet}. GenerativePointNet~\cite{xie2021generative} learns the distribution of point clouds (not the distribution of points) by EBM for point cloud generation. ShapeGF~\cite{cai2020learning} considered a point cloud on the surface of an object to be sampled from a probability distribution. The gradient of the log-density of the smoothed probability distribution (score function) is parameterized by NN and learned using denoising score matching.

In this work, we parameterize the energy function of each EBM with a coordinate-based NN such that the object surface is a level-set and optimize each EBM using MCMC sampling.

\section{Energy-Based Models}\label{EBMexpo}
In this section, we introduce the preliminary of EBMs for the proposed method.

Given a data point $\vect{x} \in \R^d$, let $E_\theta (\vect{x})\in \R$ be an energy function evaluated at $\vect{x}$.  In this paper, this function is modeled by a neural network with parameters $\theta$. The energy function defines a probability distribution called the Gibbs distribution, 
\begin{equation}
    p_\theta(\vect{x})=\frac{\exp\left(-\beta E_\theta(\vect{x})\right)}{Z(\theta)}~, 
\end{equation}
where $Z(\theta)=\int{\exp(-\beta E_\theta(\vect{x}))d\vect{x}}$ is the partition function. $\beta$ is a hyperparameter called inverse temperature. If $\beta \to 0$, the distribution goes to be uniform, and if $\beta \to \infty$, the distribution goes to be such that only the lowest energy region has a large probability density.

\subsection{Maximum Likelihood Estimation}

Let $\mathcal{P}=\{\vect{x}_i \in \R^3\}_{i=1}^N$ be the observed point cloud. We estimate the point cloud distribution $p(\vect{x})$ by minimizing the expected value of the negative log-likelihood, 
\begin{equation}
    \mathcal{L}_{\rm ML}(\theta) = \mathbb{E}_{\vect{x} \sim \mathcal{P}}[- \log p_\theta(\vect{x})],
\end{equation}
where $-\log p_\theta(\vect{x}) = \beta (E_\theta(\vect{x}) + \log Z(\theta)).$ The gradient of this objective function is calculated~\cite{turner2005cdnotes, song2021train} as 
\begin{equation}
    \nabla_\theta \mathcal{L}_{\rm ML}(\theta) = \beta (\mathbb{E}_{\vect{x} \sim \mathcal{P}}[\nabla_\theta E_\theta(\vect{x})] - \mathbb{E}_{\vect{x} \sim p_\theta(\vect{x})}[\nabla_\theta E_\theta(\vect{x})]).
\end{equation}
Updating the parameter $\theta$ according to this gradient corresponds to reducing the energy where the point exists and increasing the energy in the whole space. To calculate this gradient, we need to sample points from the distribution $p_\theta$. In this study, we sample points using Langevin dynamics.

\subsection{Langevin dynamics}
MCMC samplings such as random walk or Gibbs sampling requires long mixing times. To avoid this problem, we use Langevin dynamics, which is used in~\cite{du2019implicitEBM} to train high-dimensional data. This method uses the gradient of the energy function to sample points by iterative updates as follows:
\begin{equation}
    x_{k+1} = x_k - \alpha_k \displaystyle \nabla_x E_\theta(x_k) + {\epsilon_k}\sqrt{2\alpha_k}~,
\end{equation}
where $\epsilon_k$ is a random number following the standard normal distribution $\mathcal{N}(0, 1)$ and the sequence $\{\alpha_k\}_{k=0}^\infty$ satisfies
\begin{equation}
    \sum_{k=0}^\infty \alpha_k =\infty ~\ \land ~\ \sum_{k=0}^\infty \alpha_k^2 <\infty~.
    \label{cond:1}
\end{equation}
When $k \to \infty$, it is known that $x_k$ follows $p_\theta(\vect{x})$~\cite{welling2011bayesian}.

In Langevin dynamics, if $\alpha_k$ satisfies Eq.~\eqref{cond:1}, the sampling points from the objective distribution can be obtained independently of the initial value $\vect{x}_0$ by taking the number of updates $K$ sufficiently large. However, the choice of a suitable initial value significantly affects the speed of convergence of the sampling. In~\cite{du2019implicitEBM}, assuming that the change of probability distribution $p(x|\theta)$ due to a single update of the parameter $\theta$ is small, the previous sampling points are stored in the sample replay buffer ${\cal B}$ and used as initial values for the next Langevin dynamics. In this case, following to~\cite{du2019implicitEBM}, a mixture of points taken from ${\cal B}$ and points following a uniform distribution near the object surface in the ratio of $95:5$ is used as the initial value of Langevin dynamics.

\section{Surface Reconstruction via EBM}\label{SRviaEBM}

\begin{figure}[t]
\begin{center}
   \includegraphics[width=0.8\linewidth]{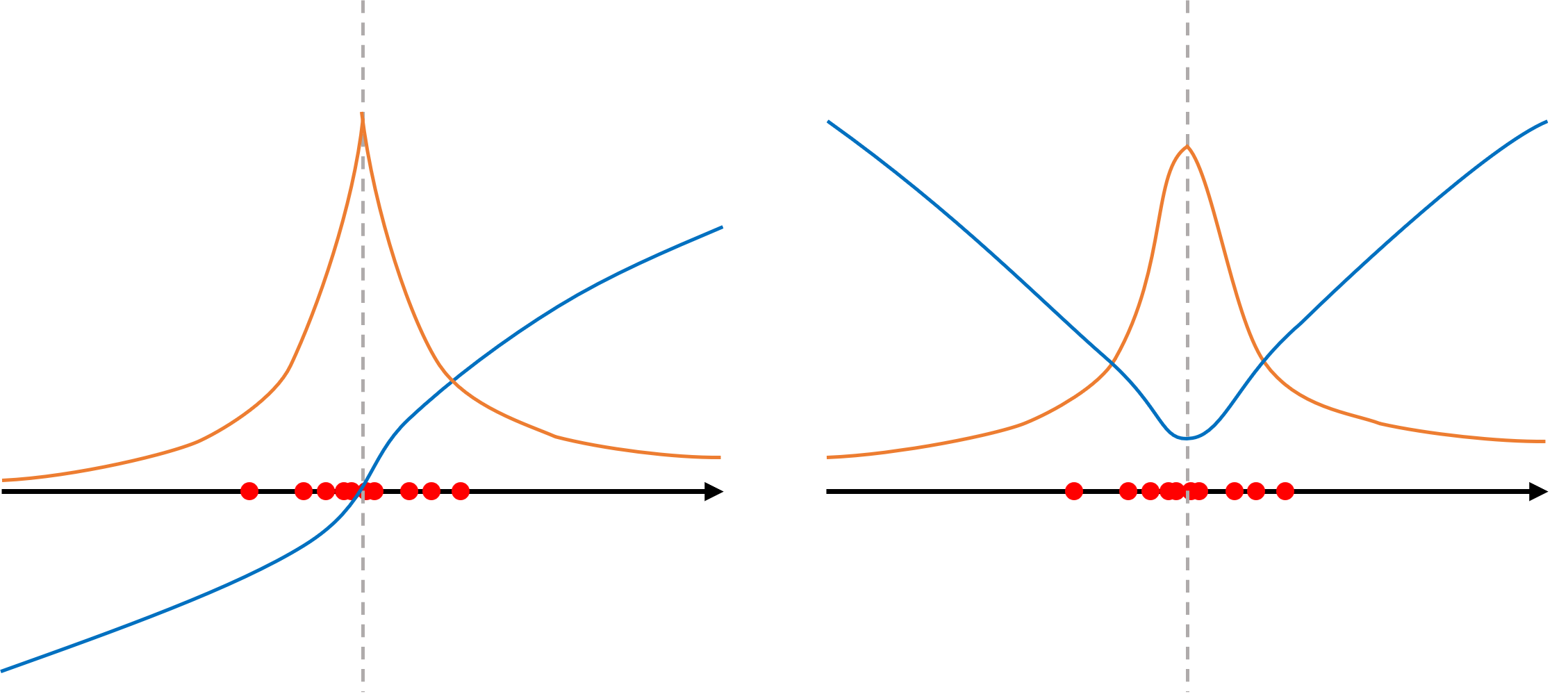}
\end{center}
   \caption{One-dimensional example showing the relationship between $f_\theta$ and $p_\theta$. The red dots are point clouds. The blue and orange lines are $f_\theta$ and $p_\theta$ derived from $f_\theta$. Left: The zero level set of $f_\theta$ represents the surface. The surface reconstruction is successful. Right: $p_\theta$ represents the point cloud distribution, but the zero level set of $f_\theta$ does not match the surface. The surface reconstruction fails.}
\label{fig:f_and_p}
\end{figure}

We model the point cloud generation distribution with an EBM that uses the absolute value $|f_\theta(\vect{x})|$ of the coordinate-based NN $f_\theta: \R^3 \to \R $ as the energy function; therefore, 
\begin{equation}
p_\theta(\vect{x})=\frac{\exp\left(-\beta |f_\theta(\vect{x})|\right)}{Z(\theta)}~,
\end{equation}
where $Z(\theta)=\int{\exp(-\beta |f_\theta(\vect{x}))|d\vect{x}}$ is the partition function. The zero level set $\mathcal{S} = \{\vect{x} \in \R^3 \mid f_\theta(\vect{x})=0\}$ is the mode of the probability distribution $p_\theta(\vect{x})$ derived from the EBM. Thus, when the probability density of the point cloud distribution is equally maximum at all points on the surface, one possible solution of maximum likelihood estimation is a situation where $\mathcal{S}$ matches the surface and the sign of $f_\theta(\vect{x})$ reverses inside and outside the surface, as in the SDF. Surface reconstruction is successful when $f_\theta$ converges to such a solution (left of Fig.~\ref{fig:f_and_p}). The surface represented by the zero level set of $f_\theta$ where the sign of $f_\theta$ change can be efficiently retrieved via Marching cubes~\cite{lorensen1987marching}.  However, there is no guarantee that the maximum value of the energy function will be $0$ merely by the maximum likelihood estimation; that is, $|f_\theta(\vect{x})|>0$ may be satisfied at any points $\vect{x}$ (right of Fig.~\ref{fig:f_and_p}), which very often causes failure of surface reconstruction. To avoid such a situation as possible, we initialize $\theta$ using geometric initialization from \cite{atzmon2020sal} and start the optimization process so that $f_\theta$ approximates the SDF of a sphere. When $f_\theta(\vect{x})$ is initialized by geometric initialization, the zero level set of $f_\theta(\vect{x})$ is a closed surface. If part of the zero level set of $f_\theta(\vect{x})$ represents a mode of point cloud distribution and the rest does not, the rest will give a large probability to a no-point region, and therefore the entire zero level set must lie on the point cloud. Thus, if the object is a watertight surface, the entire point cloud will be represented by the zero level set. 

In optimization, we impose a constraint (Eikonal term) on $f_\theta(\vect{x})$ to make its gradient norm $|\nabla_{\vect{x}} f_\theta(\vect{x})|$ close to $1$. When this constraint is satisfied, the point cloud distribution $p_\theta(\vect{x})$ defined by EBM becomes a Laplace distribution with a variance of $2/\beta^2$ near the surface. Therefore, by controlling $\beta$, we can explicitly consider the scale of point cloud noise. If the scale of noise generated during point cloud acquisition is known, setting $\beta$ accordingly can improve the quality of surface reconstruction.

The actual scanned point cloud has variations in density, and the assumption that the point cloud distribution is equally maximum at all points on the surface is not strictly satisfied. However, experiments have confirmed that this is not a significant problem. If necessary, this problem can be addressed by subsampling the point cloud. If the number of points in the point cloud is small, once $f_\theta(\vect{x})$ has been optimized (in which case the zero level set of $f_\theta(\vect{x})$ does not have to be a surface representation), sampling a large number of points from $p_\theta(\vect{x})$ and then subsample it.

\subsection{Model architecture}

We use a multi-layer perceptron (MLP) as $f_\theta({\vect{x}})$. The number of layers is eight, and skip connections are implemented between the input layer and the fifth layer. This network design is based on~\cite{park2019deepsdf}.

To represent the detailed shape of the surface, positional encoding (PE) was applied to the MLP input. positional encoding is an encoding function introduced in~\cite{mildenhall2020nerf} that maps the input to a higher-dimensional space as follows:
\begin{eqnarray}
    \gamma(p)&=&(\sin(2^0 \pi p), \cos(2^0 \pi p), \nonumber\\&& ~\ \ldots, \sin(2^{L-1} \pi p), \cos(2^{L-1} \pi p)) ~,
\end{eqnarray}
where $L$ is the dimension of positional encoding, which is a hyperparameter. Positional encoding helps NNs to represent high-frequency structures. The input coordinates are concatenated with $\gamma(p)$ and then input to MLP. We found that the surface reconstruction often fails if $f_\theta(\vect{x})$ has a high representation capability in the early stage of EBM optimization. Therefore, we adopted a strategy of increasing the dimension $L$ of positional encoding as the learning progresses~\cite{lin2021barf}.

The MLP weights are initialized by geometric initialization, a method introduced in~\cite{atzmon2020sal} that initializes the INR as an approximation of the SDF of a sphere. Intuitively, it is based on the fact that, in a linear transformation, the norm of the output is proportional to the norm of the input. Hence, geometric initialization is incompatible with positional encoding and skip-connections. To avoid this problem, we set the initial values of MLP weights applied to positional encoding and skip-connections to $0$.

\subsection{Optimization}
To optimize the EBM, we use
\begin{equation}
\begin{split}
    \mathcal{L}_{\rm EBM}(\theta) = \beta (\mathbb{E}_{\vect{x} \sim \mathcal{P}}[E_\theta(\vect{x})] - \mathbb{E}_{\vect{x} \sim p_\theta(\vect{x})}[E_\theta(\vect{x})]) \\
    = \beta (\mathbb{E}_{\vect{x} \sim \mathcal{P}}[|f_\theta(\vect{x})|] - \mathbb{E}_{\vect{x} \sim p_\theta(\vect{x})}[|f_\theta(\vect{x})|])
\end{split}
\end{equation}
as the objective function. Since $\nabla_\theta \mathcal{L}_{\rm EBM}(\theta) = \nabla_\theta \mathcal{L}_{\rm ML}(\theta)$, we can optimize the EBM by minimizing this objective function. In practice, this objective function is approximated using points sampled from the point cloud $\mathcal{P}$ (positive points) and points sampled from $p_\theta(x)$ (negative points).

As previously stated, we use the loss term to make the norm of the gradient of $f_\theta(\vect{x})$ closer to $1$: 
\begin{equation}
    \mathcal{L}_{\rm E}(\theta) = \frac{1}{|\mathcal{U}|}\sum_{\vect{x} \in \mathcal{U}}{(|\nabla_{\vect{x}} f_\theta(\vect{x})| - 1)^2} ~,
\end{equation}
where $\mathcal{U}$ is the set consisting of positive points and negative points. This term, called Eikonal term in~\cite{grop2020igr}, promotes $f_\theta$ to be SDF, and therefore, the value of $\beta$ in EBM is related to the scale of point cloud noise. Additionally, this term stabilizes the Langevin dynamics by mitigating the instability caused by sudden changes in the gradient of the energy function~\cite{du2019implicitEBM}.

Adding these two terms together, the following is the objective function:
\begin{equation}
    \label{lossfun}
    \mathcal{L}(\theta) = \mathcal{L}_{\rm EBM}(\theta) + \gamma \mathcal{L}_{\rm E}(\theta) ~,
\end{equation}
where $\gamma$ is a positive constant.

\section{Experiments}

\begin{table*}[tbp]
\begin{center}
\scalebox{0.95}{
\begin{tabular}{l|rrrr|rrrr|rrrr}
\hline\hline
\multirow{2}{*}{method} &
\multicolumn{4}{c|}{CD $(\times10^{-3})\downarrow$} &
\multicolumn{4}{c|}{F-score $(\%)\uparrow$}&
\multicolumn{4}{c}{NCS $(\times 10^{-2})\uparrow$} \\ \cline{2-13}
& 0.00 & 0.01 & 0.03 & 0.05 & 0.00 & 0.01 & 0.03 & 0.05 & 0.00 & 0.01 & 0.03 & 0.05 \\ 
\hline 
SPSR~\cite{kazhdan2013screened} & 11.59 & 13.70 & 17.82 & 24.77 & 77.24 & \bf{\underline{70.54}} & \bf{\underline{41.30}} & \bf{\underline{24.13}} & 90.85 & 86.72 & \bf{\underline{76.08}} & 65.81 \\
SAL~\cite{atzmon2020sal} & \bf{5.68} & \bf{\underline{7.15}} & 19.17 & 26.64 & \bf{\underline{82.25}} & 62.46 & 13.97 & 10.61 & \bf{\underline{91.38}} & \bf{\underline{89.09}} & 74.95 & \bf{\underline{70.67}} \\
Neural-PULL~\cite{ma2020neuralpull} & 11.35 & 9.07 & 17.21 & 24.61 & 77.14 & 42.13 & 19.77 & 14.56 & 90.10 & 78.17 & 63.15 & 58.82 \\
Neural splines~\cite{williams2021spline} & 11.34 & 12.70 & 43.99 & 50.63 & 73.45 & 70.03 & 23.74 & 14.28 & 87.70 & 86.82 & 70.34 & 60.41 \\
SAP~\cite{peng2021shape} & 14.15 & \bf{6.73} & \bf{12.15} & \bf{\underline{17.36}} & 79.07 & 61.74 & 28.78 & 19.43 & 87.20 & 69.46 & 55.51 & 52.85 \\
Ours & \bf{\underline{6.45}} & 7.67 & \bf{\underline{12.68}} & \bf{13.17} & \bf{84.74} & \bf{76.78} & \bf{49.45} & \bf{34.93} & \bf{91.60} & \bf{89.49} & \bf{85.12} & \bf{82.23} \\
\hline
Points2Surf~\cite{erler2020Points2Surf} & 4.73 & 5.32 & 7.19 & 9.44 & 86.53 & 81.69 & 62.16 & 43.66 & 89.67 & 87.83 & 82.11 & 74.86\\
POCO~\cite{boulch2022poco} & 4.19 & 4.73 & 7.75 & 10.24 & 88.23 & 82.17 & 59.19 & 43.08 & 93.64 & 92.14 & 87.87 & 84.62\\
DeepMLS~\cite{Liu2021MLS} & 4.60 & 7.66 & 10.47 & 16.09 & 86.28 & 48.11 & 20.36 & 14.22 & 91.19 & 82.51 & 61.49 & 55.84 \\
SAP~\cite{peng2021shape} & 9.58 & 10.47 & 14.10 & 17.55 & 56.56 & 47.15 & 27.98 & 20.44 & 85.67 & 84.92 & 82.10 & 79.14\\
\hline\hline
\end{tabular}
}
\end{center}

\caption{Comparison of surface reconstruction errors with three evaluation metrics on the ABC dataset.
We report quantitative results for the Gaussian scan noise at four levels ($\sigma =0.00, 0.01, 0.03, 0.05$) for 10 scans point clouds.  
The {\bf{best}} and {\bf{\underline{second best}}} non-data-driven methods are highlighted in each column.
}
\label{tab:result2}
\end{table*}

\begin{table*}[tbp]
\begin{center}
\scalebox{0.95}{
\begin{tabular}{l|rrrr|rrrr|rrrr}
\hline\hline
\multirow{2}{*}{method} &
\multicolumn{4}{c|}{CD $(\times10^{-3})\downarrow$} &
\multicolumn{4}{c|}{F-score $(\%)\uparrow$}&
\multicolumn{4}{c}{NCS $(\times 10^{-2})\uparrow$} \\ \cline{2-13}
& 0.00 & 0.01 & 0.03 & 0.05 & 0.00 & 0.01 & 0.03 & 0.05 & 0.00 & 0.01 & 0.03 & 0.05 \\ 
\hline 
SPSR~\cite{kazhdan2013screened} & 8.45 & 13.44 & 14.51 & 21.35 & 82.86 & \bf{\underline{71.44}} & \bf{\underline{40.54}} & \bf{\underline{23.40}} & 90.16 & 84.34 & \bf{\underline{74.43}} & 64.87 \\
SAL~\cite{atzmon2020sal} & 4.15 & 5.92 & 18.47 & 26.50 & 84.21 & 60.00 & 14.55 & 10.36 & 89.91 & \bf{\underline{86.30}} & 71.85 & \bf{\underline{68.15}} \\
Neural-PULL~\cite{ma2020neuralpull} & 10.48 & 7.41 & 16.94 & 25.51 & 83.69 & 43.83 & 20.25 & 14.24 & 90.12 & 75.98 & 60.88 & 57.53 \\
Neural splines~\cite{williams2021spline} & 10.34 & 18.21 & 43.29 & 54.12 & 79.46 & 66.15 & 23.32 & 13.08 & 85.97 & 82.30 & 66.66 & 59.53 \\
SAP~\cite{peng2021shape} & \bf{2.64} & \bf{\underline{5.30}} & \bf{\underline{11.15}} & \bf{\underline{17.09}} & \bf{93.60} & 64.68 & 29.22 & 19.27 & \bf{92.63} & 69.29 & 55.36 & 52.57 \\
Ours & \bf{\underline{3.80}} & \bf{5.18} & \bf{8.44} & \bf{10.79} & \bf{\underline{89.58}} & \bf{78.56} & \bf{47.30} & \bf{34.09} & \bf{\underline{91.72}} & \bf{88.64} & \bf{84.26} & \bf{81.12} \\
\hline
Points2Surf~\cite{erler2020Points2Surf} & 3.38 & 4.54 & 6.90 & 9.92 & 86.61 & 78.27 & 53.55 & 37.15 & 88.14 & 85.35 & 77.98 & 70.85 \\
POCO~\cite{boulch2022poco} & 3.46 & 4.68 & 8.50 & 10.89 & 90.29 & 79.31 & 50.99 & 37.27 & 92.28 & 89.27 & 84.42 & 80.75 \\
DeepMLS~\cite{Liu2021MLS} & 8.09 & 8.13 & 9.79 & 15.72 & 89.16 & 46.41 & 20.21 & 13.89 & 90.06 & 81.14 & 61.02 & 55.64 \\
SAP~\cite{peng2021shape}  & 7.98 & 9.08 & 13.20 & 17.05 & 53.94 & 44.72 & 27.46 & 20.19 & 83.90 & 82.69 & 78.80 & 75.53\\
\hline\hline
\end{tabular}
}
\end{center}

\caption{Comparison of surface reconstruction errors with three evaluation metrics on the FAMOUS dataset.
We report quantitative results for the Gaussian scan noise at four levels ($\sigma =0.00, 0.01, 0.03, 0.05$) for 10 scans point clouds.  
The {\bf{best}} and {\bf{\underline{second best}}} non-data-driven methods are highlighted in each column.
}
\label{tab:result1}
\end{table*}

\begin{table*}[tbp]
\begin{center}
\scalebox{0.95}{
\begin{tabular}{l|rrr|rrr|rrr}
\hline\hline
\multirow{2}{*}{method} &
\multicolumn{3}{c|}{CD $(\times10^{-3})\downarrow$} &
\multicolumn{3}{c|}{F-score $(\%)\uparrow$}&
\multicolumn{3}{c}{NCS $(\times 10^{-2})\uparrow$} \\ \cline{2-10}
& 0.00 & 0.03 & @train & 0.00 & 0.03 & @train & 0.00 & 0.03 & @train \\ 
\hline 
Ours & 4.89 & 8.49 & 7.97 & \bf{\underline{91.73}} & 49.06 & 88.76 & \bf{\underline{95.25}} & \bf{\underline{88.35}} & \bf{\underline{94.17}} \\
Points2Surf~\cite{erler2020Points2Surf} & \bf{\underline{3.47}} & \bf{\underline{6.39}} & \bf{\underline{2.66}} & 90.99 & \bf{\underline{57.95}} & \bf{\underline{94.95}} & 92.86 & 81.06 & 93.43  \\
POCO~\cite{boulch2022poco} & \bf{2.67} & \bf{6.14} & \bf{2.30} & \bf{94.55} & \bf{59.35} & \bf{96.40} & \bf{96.19} & \bf{90.84} & \bf{96.21} \\
DeepMLS~\cite{Liu2021MLS} & 4.24 & 16.76 & 3.34 & 82.34 & 41.91 & 89.43 & 92.62 & 86.60 & 93.91 \\
SAP~\cite{peng2021shape} & 5.42 & 9.33 & 3.95 & 74.18 & 37.57 & 83.58 & 91.79 & 87.69 & 93.71 \\
\hline\hline
\end{tabular}
}
\end{center}

\caption{Comparison of surface reconstruction errors with three evaluation metrics on the ShapeNet dataset. data-driven methods have been trained on the ShapeNet training categories. We report quantitative results with two noise levels ($\sigma=0.00, 0.03$) in the evaluation categories and quantitative results without noise in the training categories.
The {\bf{best}} and {\bf{\underline{second best}}} methods are highlighted in each column.
}
\label{tab:result3}
\end{table*}

\begin{figure*}[tbp]
\begin{center}
\includegraphics[width=1.0 \linewidth]{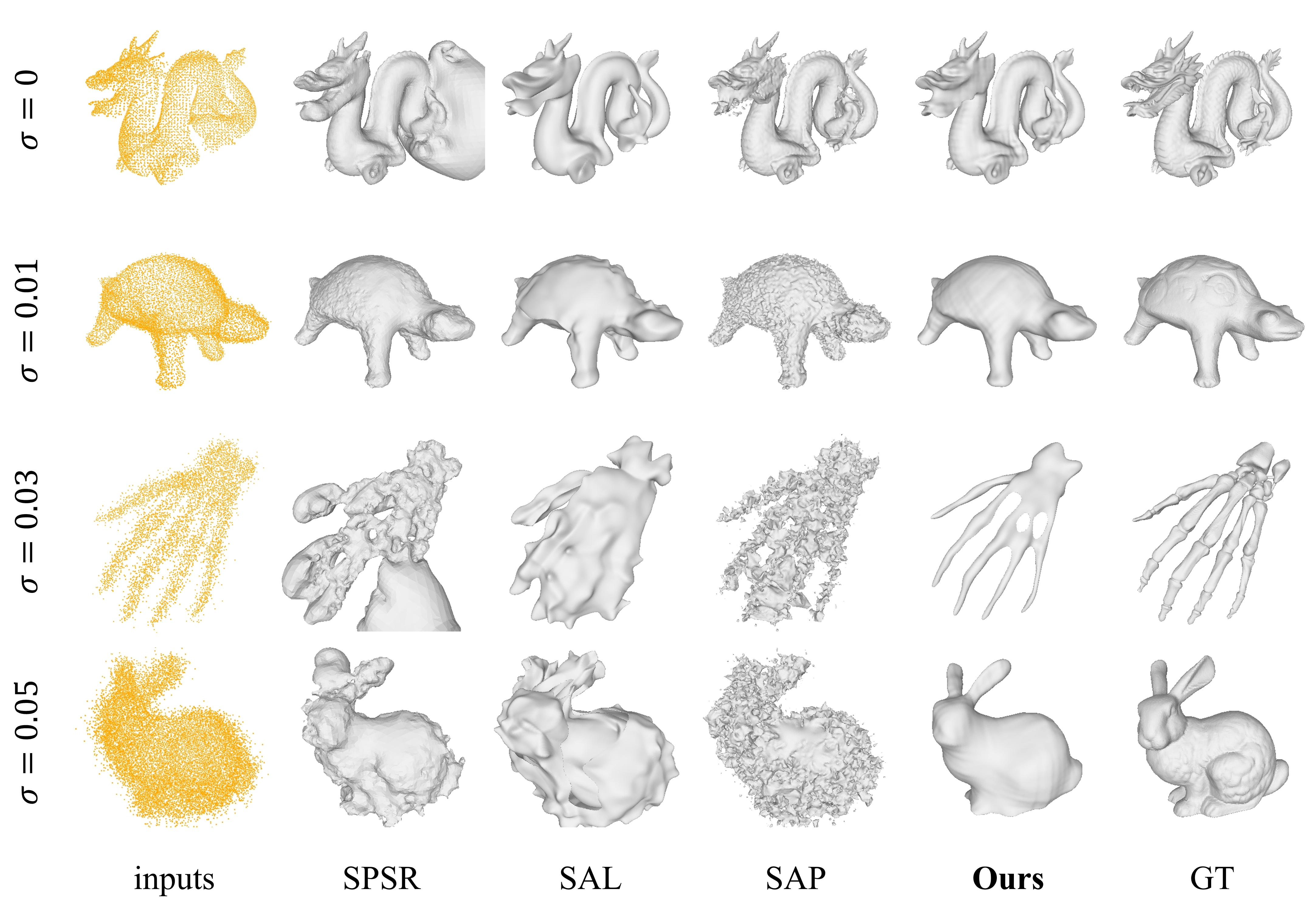}
\end{center}
   \caption{Qualitative comparison of surface reconstructions on FAMOUS dataset.}
\label{fig:qualitativefull}
\end{figure*}

\begin{figure}[t]
\begin{center}
   \includegraphics[width=0.9\linewidth]{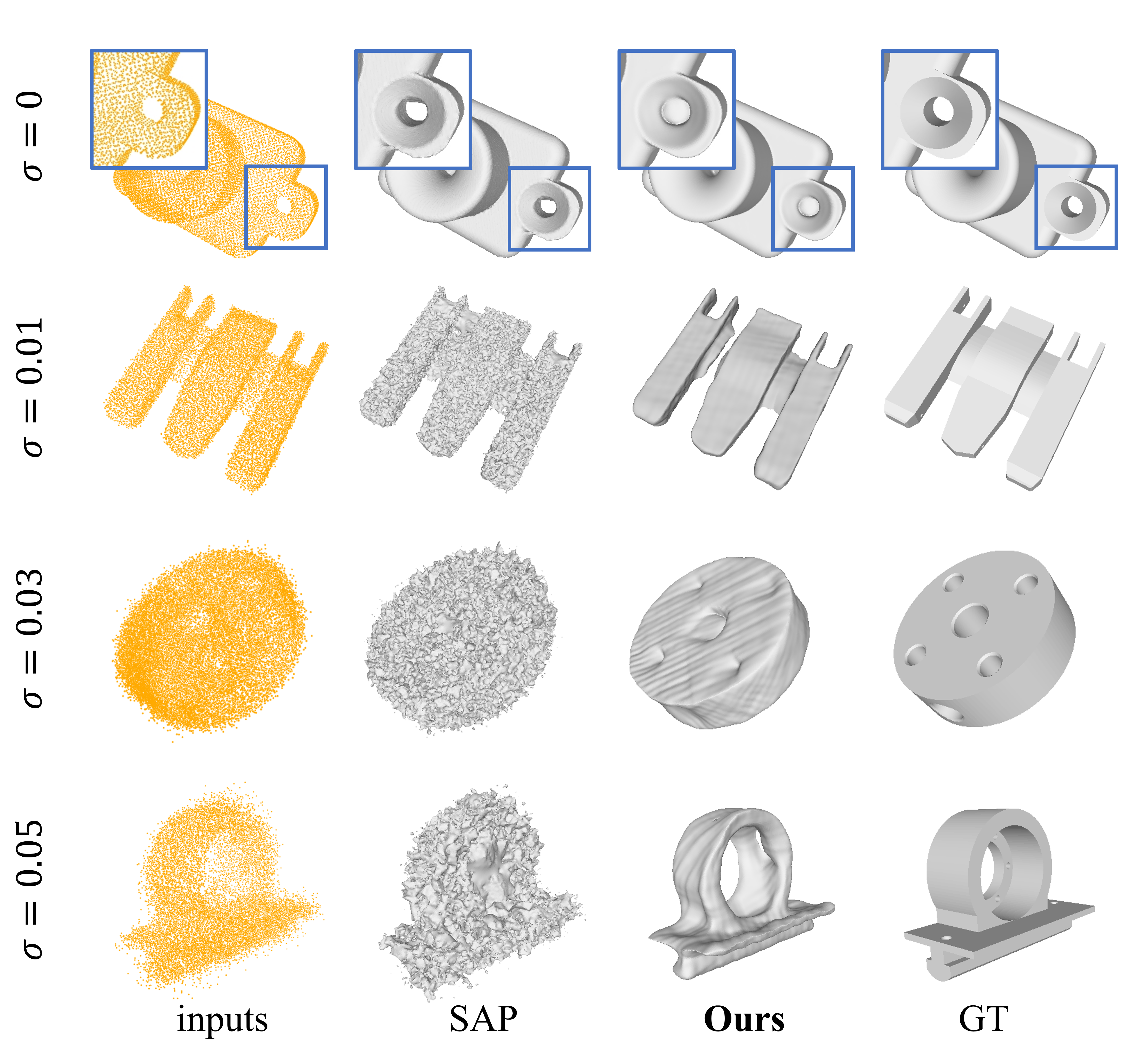}
\end{center}
   \caption{Qualitative comparison of surface reconstructions on ABC dataset.}
\label{fig:resultABC}
\end{figure}

\label{experiments}
We conducted two experiments. The first is a comparison of the proposed method with other non-data-driven methods. Second, we compared the proposed method with data-driven methods. Data-driven methods are expected to achieve high reconstruction accuracy when training data similar to the evaluation data are available. Therefore, we compared the accuracy of the proposed method with that of the data-driven methods when there is a difference between the training data and the evaluation data. If the proposed method is comparable or superior to data-driven methods under this condition, it indicates the effectiveness of the proposed method in terms of versatility and usability of the method.

\subsection{Datasets}\label{datasets}
For comparison with the non-data-driven method, we used two datasets, the ABC dataset~\cite{koch2019abc} and the FAMOUS dataset~\cite{erler2020Points2Surf}. The ABC dataset is a dataset containing a variety of CAD meshes. We used 100 meshes for evaluation and 4950 meshes for training data-driven methods. The FAMOUS dataset is a dataset containing 22 meshes well-known in geometry processing. These are the datasets used for evaluation in Points2Surf~\cite{erler2020Points2Surf}, and the selection of evaluation sets follows that of the Points2Surf paper.

For comparison with data-driven methods, we used ShapeNet dataset~\cite{chang2015shapenet}. To evaluate the performance of the data-driven method on out-of-distribution data, we divided each category into training and evaluation categories. Specifically, five categories \textit{airplane}, \textit{lamp}, \textit{chair}, \textit{table}, \textit{display} were used as evaluation categories, and eight categories \textit{sofa}, \textit{phone}, \textit{vessel}, \textit{speaker}, \textit{cabinet}, \textit{car}, \textit{bench}, \textit{rifle} were used as training categories. In addition, the evaluation set was created by referring to the train-test split of occupancy networks~\cite{mescheder2019occupancy} from among the evaluation categories.

To create a point cloud for evaluation, we sampled points from these mesh data using the Points2Surf preprocessing script. Following Points2Surf, we performed multiple scans for each mesh and merged the results into a single point cloud data set. To verify the robustness of our method to noise, we add scan noise per ray to point cloud. The noise is Gaussian noise in the depth direction, and we created four datasets with $\sigma=0,\ 0.01,\ 0.03,\ 0.05$. We set the number of scans to 10, and also created additional data sets of 5 scans (sparse) and 30 scans (dense).

\subsection{Baselines}
As a baseline for non-data-driven methods, we selected SPSR\cite{kazhdan2006poisson}, Neural Splines\cite{williams2021spline}, Neural-Pull\cite{ma2020neuralpull}, SAP\cite{peng2021shape}, and SAL\cite{atzmon2020sal}. SPSR is a classical surface reconstruction method, and we utilized its implementation in MeshLab. Although Neural-Pull, SAP, and SAL can also be used as data-driven methods, we evaluated only their geometric priors without pre-training them in order to compare them with our proposed method under the same conditions. For Neural Splines, Neural-Pull, and SAP, we used the implementations provided by the authors. As for the SAL, we reimplemented it and did not use the authors' implementation. To focus on the differences in optimization methods, we used the same network architecture for SAL as our proposed method. We also added the Eikonal term to the objective function of the SAL. By comparing the results of SAL with those of our proposed method, we confirmed the effectiveness of EBMs. SPSR, SAP, and Neural Splines require normal estimation for point clouds, and we performed normal estimation using $k=40$ nearest neighbors.

As a baseline for data-driven methods, we chose POCO~\cite{boulch2022poco}, Points2Surf~\cite{erler2020Points2Surf}, DeepMLS\cite{Liu2021MLS}, and SAP~\cite{peng2021shape}. For all of these, we used the author's implementation. We trained these methods on the ShapeNet training categories defined in the previous section. We used the default training settings, but the scale of point cloud noise added during training was adapted to our evaluation set. We also trained these methods on the ABC dataset for comparison with non-data-driven methods. The training set is the same as in the Points2Surf paper.

\subsection{Evaluation Metrics}

 After rescaling to the size of the original mesh, the performance of each surface reconstruction method is evaluated by the following three metrics (Chamfer Distance~\cite{fan2017point}, F-score~\cite{knapitsch2017tanks}, Normal Consistency Score~\cite{mescheder2019occupancy}) between two point sets ${\cal{R}}$ and $\cal{G}$ sampled from the reconstructed surface and the ground-truth surface, respectively. Here, we set $|{\cal{R}}|=|{\cal{G}}|=200{\rm K}$ for each experiment. 
{\bf{Chamfer Distance}} – Chamfer Distance (CD) is a metric between two point sets. 
 CD indicates that the smaller the value is, the more accurate the surface reconstruction is.
{\bf{F-score}} – Different from CD, F-score takes the harmonic mean between the precision and recall, which means that it will be dominated by the minimum of either precision or recall. 
Therefore, F-score is more sensitive to extreme cases in which either precision or recall is worse. 
We set the distance threshold $\tau=0.005$. 
{\bf{Normal Consistency Score}} – Normal Consistency Score (NCS) is a metric measuring the consistency between the normal vector fields of two surfaces. As such, it measures the difference of higher-order information between different surfaces and is more sensitive to subtle shape differences. 

\subsection{Implimentation details and parameter settings}
In our experiments, we set the dimension of the hidden layer of $f_\theta(\vect{x})$ to 512 and the dimension of positional encoding $L$ to 6. We set $\gamma=5.0$ in Eq.~\eqref{lossfun}. The sequence $(\alpha_k)_{k=1}^K$ was chosen as $\alpha_k = \alpha_0 / (1 + s(k-1))$, where $\alpha_0$, $s$, $\beta$, and the length of the sequence $K$ are parameters. We set $\alpha_0 = 0.03 / \beta$, $s=1$, and $K=30$ . However, we set $s=0.1$ and $K=1000$ to prevent instability at the beginning of learning and when changing $\beta$ and learning rate. These parameters are set to the same values for all experiments. %

We changed $\beta$ according to the scale of the point cloud noise. Specifically, we set $\beta= 800,\ 150,\ 50,\ 30$ for $\sigma=0,\ 0.01,\ 0.03,\ 0.05$, respectively. These values were determined so that the variance of the distribution of EBMs roughly matches the variance of the noise ($\beta \sim \sqrt{2}/\sigma$). 
Since the scale of the realistic noise derived from the performance of the scanner and the properties of the object surface can be roughly estimated by conducting preliminary experiments, we do not consider it unfair to determine $\beta$ in this way. 
We initially set $\beta$ to a small value ($\beta_{\rm{init}}=10$) and increased it as the learning progressed since using a large $\beta$ from the beginning often produced undesirable results such as the formation of surfaces in regions where no points exist or no surfaces in regions where points exist. 

We use the Adam optimizer with an initial learning rate of 0.0003, and optimize the EBM in 150 epochs. In one update, 256 positive points and 2048 negative points are used to compute the objective function.

We use the Marching cubes~\cite{lorensen1987marching} to extract a zero level set of optimized $f_\theta(\vect{x})$ as a mesh.

\begin{figure}[t]
\begin{center}
   \includegraphics[width=0.9\linewidth]{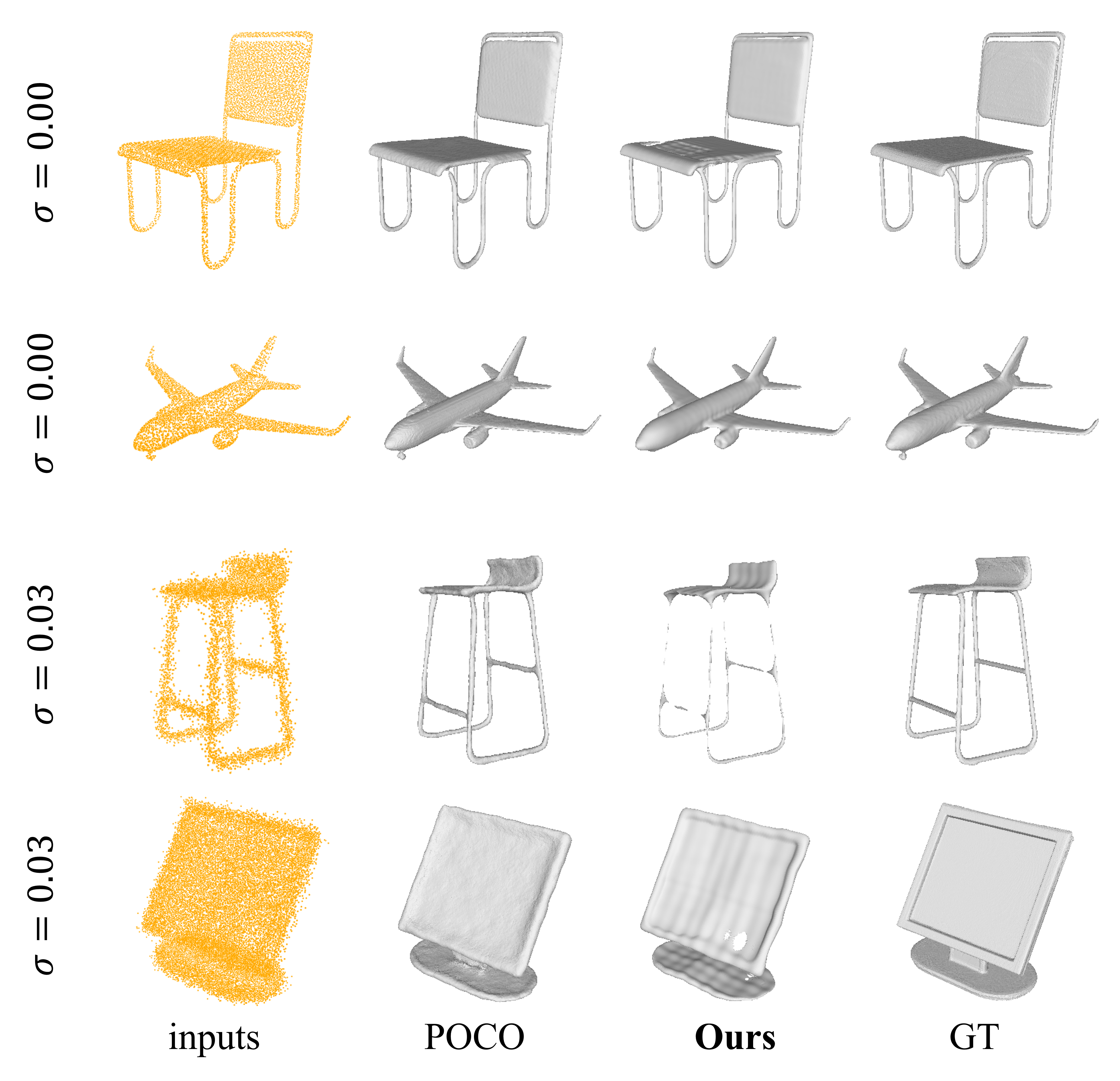}
\end{center}
   \caption{Qualitative comparison of surface reconstructions on ShapeNet dataset.}
\label{fig:resultShapeNet}
\end{figure}

\section{Results}
\label{results}
\subsection{Comparison with non-data-driven methods}
Table~\ref{tab:result2}, Table~\ref{tab:result1} show the comparison results with the non-data-driven baseline methods on the ABC and the FAMOUS datasets. 
For reference, we also include the results of the data-driven methods (point2surf, POCO, DeepMLS, and SAP) trained on the ABC dataset in the tables.
Figure~\ref{fig:qualitativefull} shows the surfaces reconstructed by each method. Since our method and SAL use the same NN architecture as $f_\theta(\vect{x})$, we use them to investigate how noise tolerance differs between different optimization methods for non-data-driven INRs. SAL achieves comparable reconstruction accuracy as the proposed method for noiseless point clouds, but its score drops rapidly as the noise scale increases. In contrast, our method's score deteriorates much more slowly. Therefore, we can conclude that the robustness of our method to noise lies in the optimization method adopted in this study. The proposed method is comparable to the best-performing existing methods on noiseless point clouds and outperforms other methods on noisy point clouds. In particular, on the FAMOUS dataset, the proposed method achieves performance comparable to the best data-driven method. The difference between the ABC and FAMOUS datasets may have limited the performance of the data-driven methods, since the non-data-driven methods were trained only on the ABC dataset. The proposed method has a relatively poor Chamfer-L2 score on the ABC. As shown in Figure~\ref{fig:resultABC}, the ABC dataset contains many meshes with small holes, and the proposed method tends to fill these holes.

\subsection{Comparison with data-driven methods}
Table~\ref{tab:result3} presents the surface reconstruction performance of the proposed method and data-driven methods on the ShapeNet dataset. The data-driven methods were trained on the categories described in Section \ref{datasets} and evaluated on data sampled from both in-distribution and out-of-distribution categories. In particular, the out-of-distribution data was evaluated with two noise levels ($\sigma=0.00, 0.03$), while no noise was added to the in-distribution data. The data-driven methods outperform the proposed non-data-driven method on the in-distribution data. However, the difference is smaller and sometimes reversed for the out-of-distribution data. For $\sigma=0.03$, the proposed method outperforms DeepMLS and SAP, which process the input point cloud globally, leading to difficulties in dealing with unknown shapes. In contrast, POCO and Points2Surf, which process the input point cloud mainly at the local level, exhibit less loss in accuracy for unknown shapes.
Figure~\ref{fig:resultShapeNet} displays the reconstructed surfaces from the ShapeNet dataset, which consists of meshes of artifact shapes, including basic shapes such as cylinders and plates that appear multiple times. For methods such as POCO and Points2Surf, which focus on local shapes, the differences between categories may not have been a significant issue.

\section{Conclusion}
\label{conclusion}
We have proposed a method to optimize INR from point clouds using EBM by estimating the point cloud distribution. Our method can account for noise scales in the distribution estimation and is thus highly noise-tolerant despite being a non-data-driven method.

An interesting future direction is to enable the EBM to represent the density of points in the surface direction since the probability density function is assumed to be constant on the surface in this study. If this assumption can be removed, it will be possible to reconstruct the surface for more realistic scanning.

\clearpage

{\small
\bibliographystyle{ieee_fullname}
\bibliography{ebsr}
}

\end{document}